\documentclass{ifacconf}

\usepackage{graphicx}      
\usepackage{natbib}        
\usepackage{comment}
\usepackage{soul}
\usepackage{amssymb}
\usepackage{color,soul}
\usepackage{aliascnt}
\usepackage{amsmath}
\usepackage{textgreek}
\usepackage{caption}
\usepackage{siunitx}
\usepackage{multirow}
\usepackage{adjustbox}
\usepackage{xcolor}
\usepackage{float}
\usepackage{subfigure}
\usepackage{caption}
\restylefloat{table}

\newaliascnt{eqfloat}{equation}
\newfloat{eqfloat}{h}{eqflts}
\floatname{eqfloat}{Equation}
\newcommand*{\ORGeqfloat}{}
\let\ORGeqfloat\eqfloat
\def\eqfloat{%
  \let\ORIGINALcaption\caption
  \def\caption{%
    \addtocounter{equation}{-1}%
    \ORIGINALcaption
  }%
  \ORGeqfloat
}
\begin{document}
\begin{frontmatter}

\title{Robust Anomaly Map Assisted Multiple Defect Detection with Supervised Classification Techniques \thanksref{footnoteinfo}} 

\thanks[footnoteinfo]{This work was supported by the Slovenian Research Agency and the European Union's Horizon 2020 program projects FACTLOG and STAR under grant agreements numbers H2020-869951 and H2020-956573.}

\author[1,2]{Jo\v{z}e M. Ro\v{z}anec}
\author[1,2]{Patrik Zajec} 
\author[3]{Spyros Theodoropoulos}
\author[4]{Erik Koehorst}
\author[5]{Bla\v{z} Fortuna}
\author[2]{Dunja Mladeni\'{c}}

\address[1]{Jo\v{z}ef Stefan Institute, Ljubljana, Slovenia, (e-mail: joze.rozanec@ijs.si, patrik.zajec@ijs.si)}
\address[2]{Jo\v{z}ef Stefan International Postgraduate School, Ljubljana, Slovenia, (e-mail: joze.rozanec@ijs.si, patrik.zajec@ijs.si)}
\address[3]{School of Electrical and Computer Engineering, National Technical University of Athens, Athens, Greece (e-mail: stheodoropoulos@mail.ntua.gr).}
\address[4]{Philips Consumer Lifestyle BV, Drachten, The Netherlands (e-mail: erik.koehorst@philips.com)}
\address[5]{Qlector d.o.o., Ljubljana, Slovenia (e-mail: blaz.fortuna@qlector.com)}

\begin{abstract}
Industry 4.0 aims to optimize the manufacturing environment by leveraging new technological advances, such as new sensing capabilities and artificial intelligence. The DRAEM technique has shown state-of-the-art performance for unsupervised classification. The ability to create anomaly maps highlighting areas where defects probably lie can be leveraged to provide cues to supervised classification models and enhance their performance. Our research shows that the best performance is achieved when training a defect detection model by providing an image and the corresponding anomaly map as input. Furthermore, such a setting provides consistent performance when framing the defect detection as a binary or multiclass classification problem and is not affected by class balancing policies. We performed the experiments on three datasets with real-world data provided by  \textit{Philips Consumer Lifestyle BV}.
\end{abstract}


\end{frontmatter}

\section{Introduction}
Increasing globalization, the need for mass customization, and competitive business environments drive faster delivery times and more efficient manufacturing processes while meeting high-quality standards (\cite{zheng2021applications}). The Industry 4.0 paradigm envisions meeting such needs by applying novel technologies (e.g., additive manufacturing, Internet of Things, virtual reality, artificial intelligence, among others) to the manufacturing domain (\cite{wichmann2019direction}).

Quality control is crucial in manufacturing companies, ensuring the products meet specific requirements and specifications. While human inspectors have traditionally performed visual inspection, the process is increasingly automated. Low-cost, high-performance and handy vision sensors have enabled the development and deployment of automated visual inspection solutions (\cite{chow2020artificial}). Such solutions mitigate multiple drawbacks of manual visual inspection, such as operator-to-operator inconsistency and quality dependence on the employees' experience, well-being, or workers' fatigue (\cite{see2012visual}). Furthermore, it increases inspection speed and enables greater scalability (\cite{garvey2018framework,escobar2018machine,chouchene2020artificial}).

Among the challenges of artificial-intelligence-based solutions are (i) the ability to provide a highly precise solution that matches or supersedes human ability regarding the quality of inspection, (ii) that the solution works at least as fast as humans, and (iii) provides enough flexibility to address a variety of products. Furthermore, it is desired that the pieces considered defective are regularly manually inspected (\cite{ren2021state}). This provides means to improve the machine learning models further. Some degree of models' explainability (\cite{meister2021explainability}) and defect hinting can be desired to ensure manual inspection and data labeling are performed efficiently (\cite{rovzanec2022towards,rovzanecenhancing}).

To address the challenges described above, we conducted a series of experiments comparing how machine learning models' performance varies in three scenarios: (a) when trained with the sensed images, (b) when trained with an anomaly map, and (c) when trained with an image and anomaly map. Furthermore, we artificially generate greater data imbalance to understand whether the models degrade upon greater imbalance. 
The machine learning models were developed and tested with three datasets of images provided by the \textit{Philips Consumer Lifestyle BV} corporation.

The rest of this paper is structured as follows: Section~\ref{S:RELATED-WORK} presents related work,  Section~\ref{S:USE-CASE} describes the use case on which we conducted the research, Section~\ref{S:EXPERIMENTS} describes the methodology we followed and the experiments we performed, and Section~\ref{S:RESULTS} presents the results we obtained, and their implications. Finally, in Section~\ref{S:CONCLUSION}, we provide our conclusions and outline future work.

\section{Related Work}\label{S:RELATED-WORK}
Product inspection is an important step in the production process, given product quality is one of the most critical factors in increasing revenue and retaining brand reputation (\cite{xu2018knowledge}). While historically, in many cases, it represented the largest single cost in manufacturing, many techniques for automated visual inspection have been introduced to alleviate costs and other issues (\cite{chin1982automated,czimmermann2020visual}). Automated visual inspection is being increasingly introduced in manufacturing settings, benefiting from the decreased cost of sensors and the use of artificial intelligence (\cite{benbarrad2021intelligent,peres2020industrial}).

Many approaches have been proposed to automate visual inspection. Many methods developed in the early days (e.g., compare an image, a mask or CAD model of a good product vs. a defective one and highlight potential errors by subtracting the images (\cite{chin1982automated,borish2019defect})) have evolved with the advancements of artificial intelligence (e.g., use of autoencoders to learn a non-defective component, and later highlight differences observed when encoding the image of a defective piece (\cite{zavrtanik2021draem}), or by using siamese networks (\cite{deshpande2020one})). Furthermore, while defect detection undoubtedly provides value, defect classification poses additional challenges (e.g., data collection, labeling, and supervised models' training \cite{shirvaikar2006trends}) but also enables root-cause analysis to solve quality problems at their root and avoid their recurrence (\cite{xu2018knowledge}). Current state-of-the-art image processing techniques involve deep learning, either as pre-trained models, feature extractors or for end-to-end learning (\cite{bovzivc2021mixed,pouyanfar2018survey,long2015fully,glasmachers2017limits}). While anomaly maps have been extensively used to inform humans where defects may be located (\cite{piciarelli2018vision,chow2020anomaly,tao2022unsupervised,zavrtanik2021draem}), we found few scientific papers reporting on leveraging them as an additional features' source for machine learning models (e.g., \cite{chow2020artificial}).

\section{Use Case}\label{S:USE-CASE}
\begin{figure}
\begin{center}
\includegraphics[width=8.4cm]{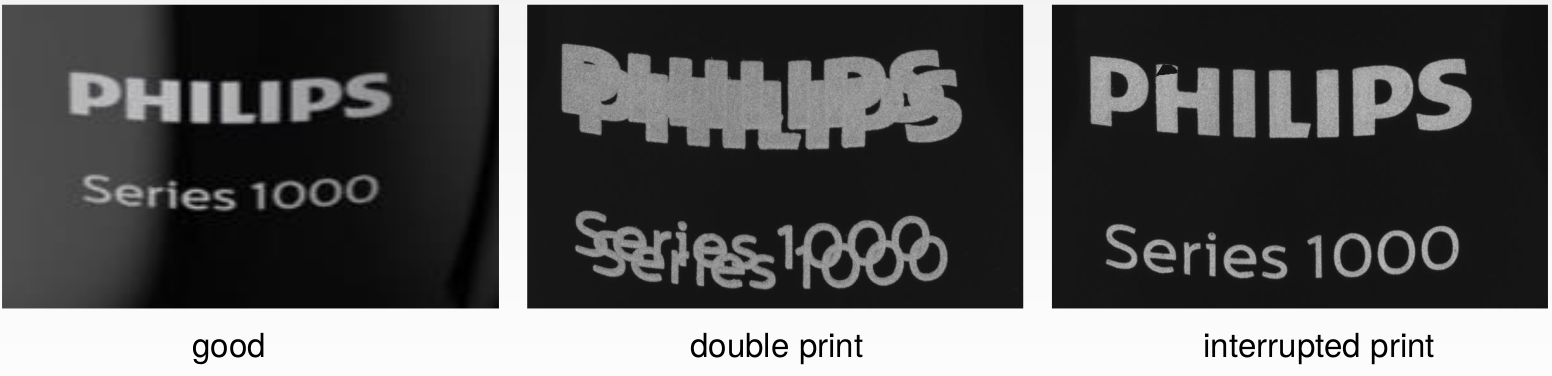}
\caption{Sample of the Shavers dataset.} 
\label{F:SHAVERS}
\end{center}
\end{figure}

\begin{figure}
\begin{center}
\includegraphics[width=8.4cm]{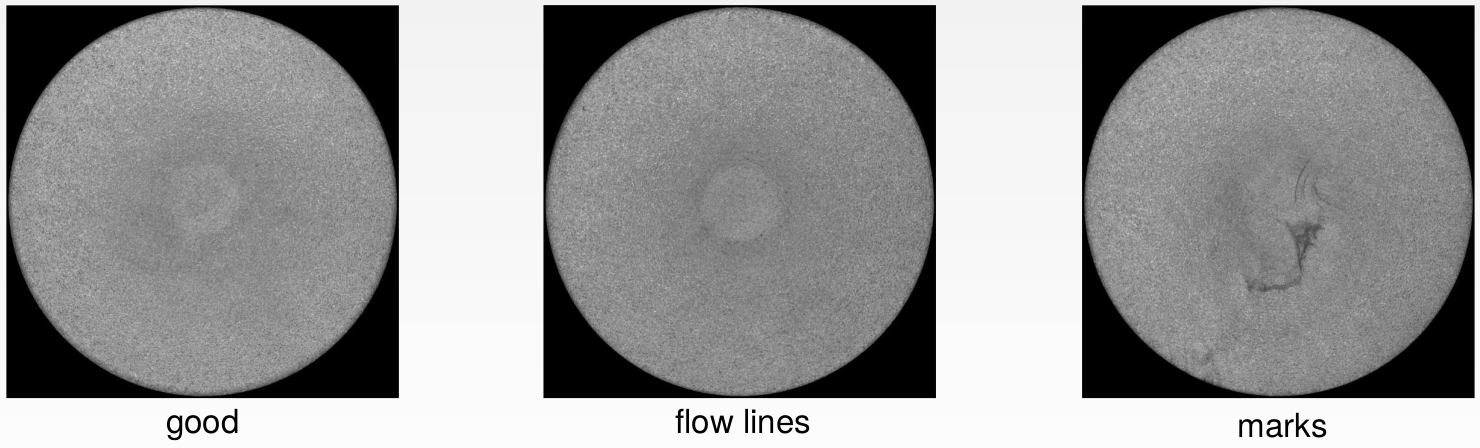}
\caption{Sample of the Deco cap dataset.} 
\label{F:DECOCAP}
\end{center}
\end{figure}

\begin{figure}
\begin{center}
\includegraphics[width=8.4cm]{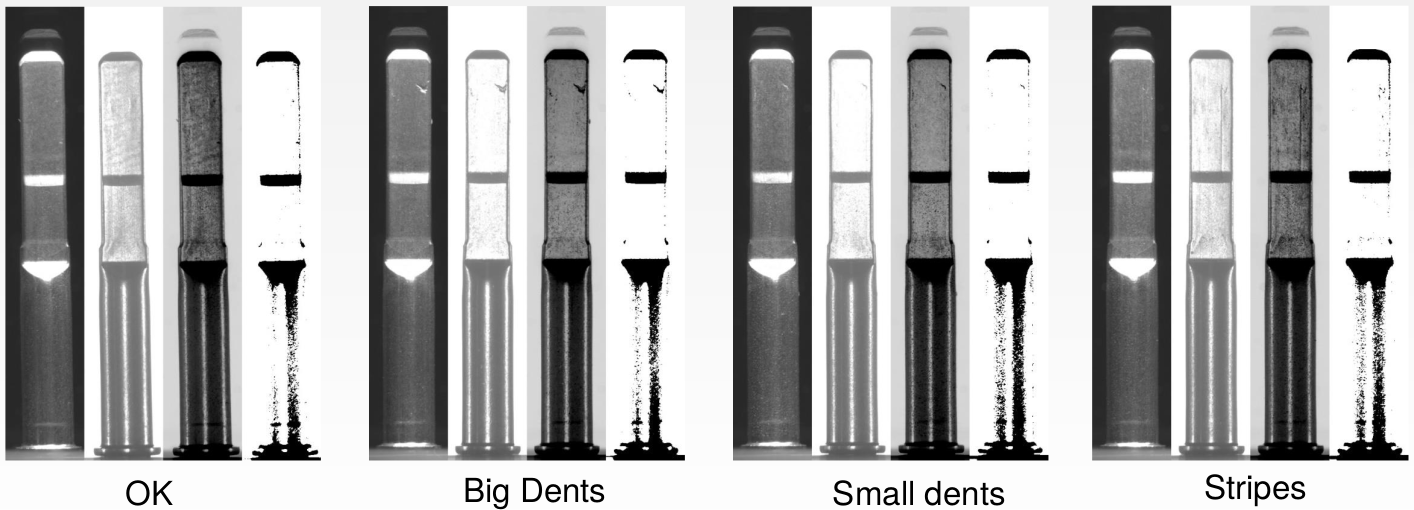}
\caption{Sample of the Shaft dataset.} 
\label{F:SHAFT}
\end{center}
\end{figure}

We performed our research on three real-world datasets of images provided by \textit{Philips Consumer Lifestyle BV} from Drachten, The Netherlands. The manufacturing plant is considered one of Europe's largest Philips development and production centers. The three datasets concern different products (shavers' logo print (3,518 images), deco cap (592 images), and shaft (4,249 images)) on which manual visual inspection was performed. The deco cap covers the center of the metal shaving head and leaves room for a print  to identify it from other types. The shaft is the toothbrush part that transfers the motion from the handle to the actual brush. The operators who perform a manual visual inspection spend several seconds handling and inspecting the product to decide whether it is defective. For each product, different defects were identified and labeled (see Fig. \ref{F:SHAVERS}, \ref{F:DECOCAP}, and \ref{F:SHAFT}). Furthermore, different degrees of class imbalance was observed for each of them (see Table \ref{T:DATASET-DESCRIPTION}).

\begin{table}[ht]
\begin{center}
\begin{tabular}{ll|c}
Dataset & Class & Number of examples \\ \hline
Deco cap & flowlines & 198 \\
 & good & 203 \\
 & marks & 191 \\ \hline
Shaft & big & 2616 \\
& good & 528 \\
& small & 954 \\
& stripe & 151 \\ \hline
Shavers & double & 244 \\
& good &  2676 \\
& interrupted & 598 \\ \hline
\end{tabular}
\end{center}
\caption{Datasets description, describing label types and the number of data instances per label for each dataset.}
\label{T:DATASET-DESCRIPTION}
\end{table}

\section{Methodology and Experiments}\label{S:EXPERIMENTS}

In this research, we studied whether the performance of machine learning models could be enhanced, including features resulting from anomaly maps. Our intuition was that as anomaly maps highlight specific regions where defects are most likely present and help humans better decide whether a defect is present, this information could be valuable when training a machine learning model. We, therefore, studied three scenarios, training supervised machine learning models: (i) solely with the original images, (ii) only with the anomaly maps, and (iii) with features resulting from the images and anomaly maps. Furthermore, we were interested in how resilient the three types of models would be to higher class imbalance. To that end, we performed the experiments including 25\%, 50\%, 75\%, and 100\% of the images regarding defective products present in the original datasets. We consider such results relevant not only to understand the models' resiliency but also to inform data collection and labeling efforts. If fewer defective samples are required to achieve the same performance, then data collection times could be shortened, and labeling efforts reduced.

\begin{figure}
\begin{center}
\includegraphics[width=8.4cm]{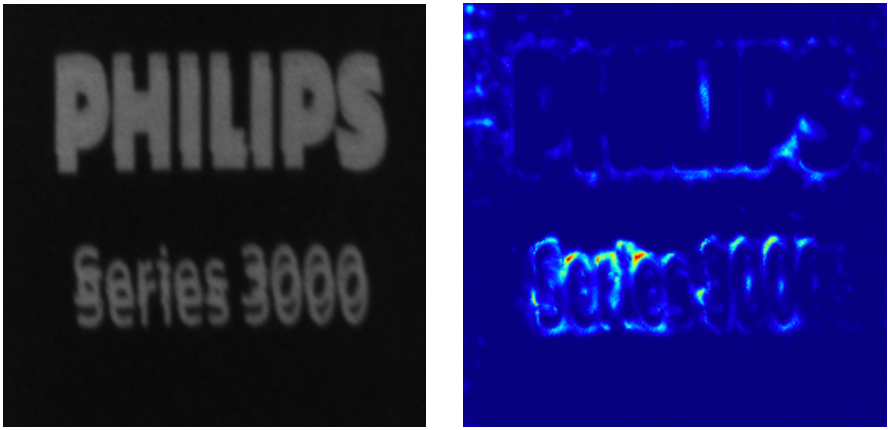}    
\caption{Image and DRAEM generated anomaly map for double print defect from Shavers dataset.} 
\label{F:IMAGE_HEATMAP}
\end{center}
\end{figure}

To create the anomaly maps, we trained a DRAEM model (\cite{zavrtanik2021draem}), a SOTA model for unsupervised defect detection. DRAEM works by training an autoencoder on good (non-defect) images. 
It aims first to reconstruct artificially corrupted images to look like non-defective pieces. The original and reconstructed images are then fed to a discriminative sub-network, which identifies the anomalous regions to create an anomaly map (see Fig. \ref{F:IMAGE_HEATMAP}).

We studied the machine learning models' performance as a binary classification problem (discriminating defective from non-defective products) and a multiclass classification problem (classifying the images into particular defect categories specific to each dataset). We measured the models' discriminative power by computing the AUC ROC metric. In the multiclass setting, we adopted the "one-vs-rest" heuristic, computing the AUC ROC metric for each class and then computing the final value as the weighted average, considering the proportion of samples of each class in the dataset. Furthermore, we also computed two types of recall for each experiment: correct defect class recall and defect recall. Correct defect class recall indicates the percentage of defective items identified as defective considering all defective instances for the particular class being analyzed. Defect recall, on the other hand, was computed as the binary recall, considering only whether the items were identified as defective considering all defective instances, regardless the class of defect. The acceptance quality level (AQL) achieved by our models can be computed through Eq. \ref{E:AQL}.

\begin{eqfloat}
\begin{equation}\label{E:AQL}
    AQL_{defect_i} = 100 - recall_{defect_i} 
\end{equation}
\caption{Equation denoting how the AQL is computed for a particular defect \textit{i}.}
\end{eqfloat}

When training the models, we performed a stratified ten-fold cross-validation (\cite{zeng2000distribution,kuhn2013applied}). We used a pre-trained ResNet-18 model to extract the features from the average pool layer. To create the anomaly maps, we trained a DRAEM model on good images for each fold, using the parameters recommended by (\cite{zavrtanik2021draem}).

To ensure a large number of features would not produce overfitting during training times, we performed feature selection, ranking them according to their mutual information and selecting the top \textit{K} features considering $K=\sqrt{N}$, with N equal to the number of data instances in the train set (\cite{hua2005optimal}).

Through the experiments, we used the multi-layer perceptron (MLP) as the classifier, consisting of two dense layers (with 512 and 100 features), with an intermediate ReLU activation between both dense layers and a softmax activation at the output. We used MLP, which proved to be a good choice based on our previous work (\cite{rovzanec2022human}).
We made the code available in a publicly accessible repository to promote research reproducibility \footnote{The repository URL will be provided upon paper acceptance. The datasets will remain confidential, as requested by \textit{Philips Consumer Lifestyle BV}.}. We present the results obtained and conclusions in Section \ref{S:RESULTS}. When assessing the results, we computed the statistical significance of the differences between the results using the Wilcoxon signed-rank test (\cite{wilcoxon}) with a p-value of $0.005$.

\section{Results}\label{S:RESULTS}
This section describes the results obtained through the experiments described in Section \ref{S:EXPERIMENTS}. 
In Table \ref{T:RECALL} we inform the recall achieved for defective pieces, considering correct defect class recall, and defect recall, while in Table \ref{T:AUC-ROC} we present results that inform the defect recall and discriminative power achieved by the defect detection models for different levels of class imbalance.

Columns under \textit{Correct defect class recall} in Table \ref{T:RECALL} show the ratio between correctly identified defects versus the number of all defects of each particular class and features.
Here, the results when using both image and anomaly map as the input are almost always on-par or better than the results achieved when using only image or anomaly map as the input. 
This demonstrates that the model benefits from additional signals introduced by the anomaly map.

The exception is the \textit{stripe} class from the \textit{Shaft} dataset, where all approaches struggle to differentiate this defect from others, with anomaly maps achieving the best score of $2.65\%$.
Note that \textit{stripe} examples are still correctly identified as defects in $95.36\%$ as achieved by the model using images and anomaly maps, they are mostly not assigned to the right class of defect.

Further, combining images and anomaly maps achieves the best results for all defect classes as seen in the columns under \textit{Defect recall} in Table \ref{T:RECALL}.
All defects are identified for the \textit{Deco cap} dataset, while scores well above $90\%$ are achieved in all other cases.
Using anomaly maps alone frequently achieves better scores than using images. This is not surprising as the anomaly map shows the difference between the current and no-defect images. Any major activity in the anomaly map can be easily recognized. Here DRAEM, which is specially tailored for surface anomaly detection, learns the notion of a no-defect image, while the model has to do it in the case of images.

Detecting defects is the primary concern of visual quality inspection and is more important than classifying the defect into the right class. 
The acceptable performance rates are usually above $99\%$ (or less than one percent AQL), which is achieved for all but three defect classes.
Here further improvements might be necessary to obtain a usable solution.

\begin{table*}[ht]
\begin{center}
\begin{tabular}{ll|ccc|ccc}
& & \multicolumn{3}{c}{Correct defect class recall} & \multicolumn{3}{c}{Defect recall} \\ 
Dataset & Class & Image & Anom. map & Image + Anom. map & Image & Anom. map & Image + Anom. map \\ \hline
Deco cap & flowlines &     98.99\% &      98.99\% &             98.99\% &       98.99\% &     100.00\% &            100.00\% \\
        & marks &     \textbf{99.48\%} &      97.38\% &             98.43\% &       99.48\% &     100.00\% &            100.00\% \\ \hline
Shaft & big &     \textbf{92.43\%} &      87.27\% &             92.24\% &       99.73\% &      98.47\% &             \textbf{99.96\%} \\
        & small &     59.85\% &      47.38\% &             \textbf{68.03\%} &       78.93\% &      90.46\% &             \textbf{92.77\%} \\ \hline
        & stripe &      0.00\% &      \textbf{2.65\%} &              0.66\% &       86.75\% &      94.70\% &             \textbf{95.36\%} \\
Shavers & double &     97.95\% &      88.11\% &             \textbf{98.36\%} &       98.36\% &      96.72\% &            \textbf{99.18\%} \\
        & interrupted &     84.45\% &      91.14\% &             \textbf{92.64\%} &       84.62\% &      92.64\% &             \textbf{93.14\%} \\
\end{tabular}
\end{center}
\caption{Percentage of images correctly classified as defects. We compute correct defect class recall (considers particular types of defects) and defect recall (binary recall - only considers defective/non-defective pieces). The best results, when statistically significant compared to the second-best results, are bolded.}
\label{T:RECALL}
\end{table*}

Table \ref{T:AUC-ROC} shows the classification performance on the models where a more severe imbalance of defect images is artificially introduced. 
The columns under \textit{Defect recall (Binary)} report the recall score for defect versus no defect setting, that is, the ratio of detected defect versus all defects. The columns under \textit{ROC AUC (Multiclass)} report the classification score where the defect has to be detected and assigned to the correct class.
The percentage on top of each column specifies how many defect images were kept from the original dataset, with lower values resulting in a more imbalanced dataset. For example, in the first column, only $25\%$ of defect images were included in the set, while the number of good images was not reduced.

The results for \textit{Defect recall} show that using images and anomaly maps is more robust for the class imbalance.
On \textit{Shavers} dataset, the performance drop when using only images is almost $12\%$ (absolute) when comparing the full dataset ($0.8860$ defect recall for $100\%$) with artificially imbalanced one ($0.7670$ for $25\%$). Making the same comparison, the drop in performance is only $3\%$ absolute when both images and anomaly maps are used.
Similarly, the drop is smaller when using only anomaly maps than when using only images. This suggests that the anomaly maps themselves provide a strong hint and make it easier for the model to distinguish between the good images and defects from the smaller number of examples.

The results for multiclass classification in \textit{ROC AUC (Multiclass)} show that using only images tends to outperform the anomaly maps, as the performance for images is higher or the difference is not significant.
Although anomaly maps highlight the presence of defects on the image, they appear not fine-grained enough to successfully differentiate between the categories of defects.
Combining images and anomaly maps is also the best approach in this setting.

\begin{table*}[ht]
\begin{center}
\begin{tabular}{ll|cccc|cccc}
{} & {} & \multicolumn{4}{c}{Defect recall (Binary)} & \multicolumn{4}{c}{ROC AUC (Multiclass)} \\
Dataset & Features & 25\% &     50\% &     75\% & 100\% & 25\% &     50\% &     75\% & 100\% \\ \hline
Deco cap & image &             0.9665 &  0.9922 &  0.9922 &  0.9922 &          0.9996 &  0.9997 &  0.9993 &  0.9999 \\
        & anom. map &             0.9949 &  1.0000 &  1.0000 &  1.0000 &          0.9976 &  0.9980 &  0.9984 &  0.9985 \\
        & image + anom. map &             0.9949 &  1.0000 &  1.0000 &  1.0000 &          0.9983 &  0.9988 &  0.9994 &  0.9996 \\ \hline
Shaft & image &             0.8417 &  0.9054 &  0.9430 &  0.9387 &          0.9146 &  0.9242 &  0.9293 &  0.9274 \\
        & anom. map &             0.9059 &  0.9473 &  0.9629 &  0.9626 &          0.8298 &  0.8502 &  0.8585 &  0.8627 \\
        & image + anom. map &             \textbf{0.9106} &  \textbf{0.9546} &  0.9669 &  \textbf{0.9793} &          0.9158 &  \textbf{0.9324} &  \textbf{0.9388} &  \textbf{0.9398} \\ \hline
Shavers & image &             0.7670 &  0.8631 &  0.8620 &  0.8860 &          0.9839 &  0.9881 &  0.9891 &  0.9897 \\
        & anom. map &             0.8705 &  0.9039 &  0.9263 &  0.9382 &          0.9854 &  0.9899 &  0.9910 &  0.9915 \\
        & image + anom. map &             \textbf{0.9181} &  \textbf{0.9465 }&  \textbf{0.9525} &  \textbf{0.9489} &          \textbf{0.9929} &  \textbf{0.9949} &  \textbf{0.9962} &  \textbf{0.9964} \\
\end{tabular}
\end{center}
\caption{Classification results. The best results, when statistically significant compared to the second-best results, are bolded.}
\label{T:AUC-ROC}
\end{table*}

\section{Conclusion}\label{S:CONCLUSION}
Based on the results, we conclude that improvement in the performance can be achieved by using both images and anomaly maps as the input to the model. Using anomaly maps also helps with the class imbalance and leads to a smaller performance drop than using only images when the imbalance is more severe.
Combining the proposed approach with data imbalance mitigation techniques could lead to further improvements.


\bibliography{main}             
                                                   







\end{document}